\documentclass[twoside,leqno,twocolumn]{article}
\usepackage{ltexpprt}
\usepackage{hyperref}
\usepackage{cite}
\usepackage{url}
\usepackage[tight,footnotesize]{subfigure}
\usepackage{balance}
\usepackage{multirow}
\usepackage{fancybox}
\usepackage{mathrsfs}
\usepackage{wrapfig}
\usepackage{graphicx}
\usepackage{amsmath}
\usepackage[mathscr]{euscript}
\usepackage{algorithm}
\usepackage{algorithmic}

\usepackage{amsthm}

\theoremstyle{definition}
\newtheorem{definition}{Definition}[section]
\newtheorem{example}{Example}[section]
\theoremstyle{remark}

\usepackage{ctable}
\usepackage{bbm}
\usepackage{verbatim}
\usepackage{comment}
\usepackage{caption}
\usepackage{epstopdf}
\usepackage{cancel}
\usepackage[normalem]{ulem}

\begin{document}

\title{Trust from the past: Bayesian Personalized Ranking based Link Prediction in Knowledge Graphs}

\author{Baichuan Zhang\thanks{Department of Computer and Information Science,
                Indiana University - Purdue University Indianapolis, USA, bz3@umail.iu.edu. The work was conducted during author's internship at Pacific Northwest National Laboratory} \\
	\and 
	Sutanay Choudhury\thanks{Pacific Northwest National Laboratory, Richland, WA, USA, \{Sutanay.Choudhury,Khushbu.Agarwal,Sumit.Purohit\}@pnnl.gov} \\
	\and 
	Mohammad Al Hasan\thanks{Department of Computer and Information Science,
                Indiana University - Purdue University Indianapolis, USA, \{alhasan,xning\}@cs.iupui.edu} \\
	\and 
	Xia Ning\footnotemark[3]
	\and
	Khushbu Agarwal\footnotemark[2]
	\and
	Sumit Purohit\footnotemark[2]
	\and
	Paola Pesantez Cabrera\thanks{Department of Computer Science, Washington State University, Pullman, WA, USA, p.pesantezcabrera@email.wsu.edu} 	
} 


\date{}

\maketitle

\begin{abstract} 

Link prediction, or predicting the likelihood of a link in a knowledge graph
based on its existing state is a key research task. It differs from a
traditional link prediction task in that the links in a knowledge graph are
categorized into different predicates and the link prediction performance of
different predicates in a knowledge graph generally varies widely. In this
work, we propose a latent feature embedding based link prediction model which
considers the prediction task for each predicate disjointly. To learn the model
parameters it utilizes a Bayesian personalized ranking based optimization
technique.  Experimental results on large-scale knowledge bases such as YAGO2
show that our link prediction approach achieves substantially higher
performance than several state-of-art approaches.  We also show that for a
given predicate the topological properties of the knowledge graph induced by
the given predicate edges are key indicators of the link prediction performance of
that predicate in the knowledge graph.

\end{abstract}



\section{Introduction}

A knowledge graph is a repository of information about entities, where entities
can be any thing of interest such as people, location, organization or even
scientific topics, concepts, etc. An entity is frequently characterized by its
association with other entities. As an example, capturing the knowledge about a
company involves listing its products, location and key individuals.
Similarly, knowledge about a person involves her name, date and place of birth,
affiliation with organizations, etc.  Resource Description Framework (RDF) is a
frequent choice for capturing the interactions between two entities.  A RDF
dataset is equivalent to a heterogeneous graph, where each vertex and edge can
belong to different classes.  The class information captures taxonomic
hierarchies between the type of various entities and relations. As an example,
a knowledge graph may identify Kobe Bryant as a basketball player, while its
ontology will indicate that a basketball player is a particular type
of athlete.  Thus, one will be able to query for famous athletes in the United
States and find Kobe Bryant.  

The past few years have seen a surge in research on knowledge representations
and algorithms for building knowledge graphs. For example, Google Knowledge
Vault~\cite{Dong.Lao.ea:14}, and IBM Watson~\cite{Fan.David.ea:10} are
comprehensive knowledge bases which are built in order to answer questions from
the general population. As evident from these works,
it requires multitude of
efforts to build a domain specific knowledge graph, which are, triple
extraction from nature language text, entity and relationship
mapping~\cite{Yao.McCallum.ea:13}, 
and link prediction~\cite{Nickel.Tresp.ea:15}.  Specifically, triples extracted
from the text data sources using state of the art techniques such as OpenIE
\cite{etzioni2008open} and semantic role labeling \cite{collobert2011natural}
are extremely noisy, and simply adding noisy triple facts into knowledge graph
destroys its purpose. So computational methods must be devised for deciding
which of the extracted triples are worthy of insertion into a knowledge graph.
There are several considerations for this decision making: (1) trustworthiness
of the data sources; (2) a belief value reported by a natural language
processing engine expressing its confidence in the correctness of parsing; and (3)
prior knowledge of subjects and objects. This particular work is motivated by the third factor. 

Link prediction in knowledge graph is simply a machine learning approach for
utilizing prior knowledge of subjects and objects as available in the knowledge
graph for estimating the confidence of a candidate triple.  Consider the
following example: given a social media post ``I wish Tom Cruise was the
president of United States", a natural language processing engine will extract
a triple (``Tom Cruise", ``president of", ``United States").  On the other
hand, a web crawler may find the fact that ``Tom Cruise is president of
Downtown Medical", resulting in the triple (``Tom Cruise", ``president of",
``Downtown Medical").  Although we generally do not have any information about
the trustworthiness of the sources, our prior knowledge of the entities
mentioned in this triples will enable us to decide that the first of the above
triples is possibly wrong. Link prediction provides a principles approach for
such a decision-making. Also note that, once we decide to add a triple to the
knowledge graph, it is important to have a confidence value associated with it.  

As we use a machine learning approach to compute the confidence of triple
facts, it is important that we quantitatively understand the degree of accuracy
of our prediction \cite{tan2014trust}. It is important, because for the same
knowledge graph the prediction accuracy level varies from predicate to
predicate. As an example, predicting one's school or workplace can be a much
harder task than predicting one's liking for a local restaurant. Therefore,
given two predicates ``worksAt" and ``likes", we expect to see widely varying
accuracy levels.  Also, the average accuracy levels vary widely from one
knowledge graph to another.   The desire to obtain a quantitative grasp on
prediction accuracy is complicated by a number of reasons: 1) Knowledge graphs
constructed from web text or using machine reading approaches can have a very
large number of predicates that make manual verification
difficult~\cite{Dong.Lao.ea:14}; 2) Creation of predicates, or the resultant
graph structure is strongly shaped by the ontology, and the conversion process
used to generate RDF statements from a logical record in the data.  Therefore,
same data source can be represented in very different models and this leads to
different accuracy levels for the same predicate.  3) The effectiveness of
knowledge graphs have inspired their construction from every imaginable data
source: product inventories (at retailers such as Wal-mart), online social
networks (such as Facebook), and web pages (Google's Knowledge Vault). As we
move from one data source to another, it is critical to understand what
accuracy levels we can expect from a given predicate.

In this paper, we use a link prediction~\footnote{We use link prediction and
link recommendation interchangeably.} approach for computing the confidence of
a triple from the prior knowledge about its subject and object. Many works exist
for link prediction~\cite{Hasan.Chaoji.ea:06} in social network
analysis~\cite{Chen.Hero:15}, but they differ from the link prediction in
knowledge graph; for earlier, all the links are semantically similar, but for
the latter based on the predicates the semantic of the links differs widely.
So, existing link prediction methods are not very suitable for this task. We
build our link prediction method by borrowing solutions from recommender system
research which accept a user-item matrix and for a  given user-item pair, they
return a score indicating the likelihood of the user purchasing the item.
Likewise, for a given predicate, we consider the set of subjects and objects as
a user-item matrix and produce a real-valued score to measure the confidence of
the given triple. For training the model we use Bayesian personalized
ranking (BPR) based embedding model~\cite{Rendle.Gantner.ea:09}, which has
been a major work  in the recommendation system. In addition, we also study the
performance of our proposed link prediction algorithm in terms of topological
properties of knowledge graph and present a linear regression model to reason
about its expected level of accuracy for each predicate.

Our contributions in this work are outlined below:

\begin{enumerate}

\item We implement a Link Prediction approach for estimating confidence for triples in a Knowledge Graph.  
Specifically, we borrow from successful approaches in the recommender systems domain, adopt the algorithms for knowledge graphs and perform a thorough evaluation on a prominent benchmark dataset.

\item We propose a Latent Feature Embedding based link recommendation model for prediction task and utilize Bayesian Personalized Ranking based optimization technique for learning models for each predicate (Section 4). Our experiments on the well known YAGO2 knowledge graph (Section 5) show that the BPR approach outperforms other competing approaches for a significant set of predicates  (Figure 1).

\item We apply a linear regression model to quantitatively analyze the correlation between the prediction accuracy for each predicate 
and the topological structure of the induced subgraph of the original Knowledge Graph. Our studies show that metrics such as clustering coefficient or average degree 
can be used to reason about the expected level of prediction accuracy (Section 5.3, Figure 2).

\end{enumerate}

\section{Related Work}

There is a large body of work on link prediction in knowledge graph.  In terms
of methodology, factorization based and related latent variable
models~\cite{Nickel.Tresp.ea:11,Chang.Yang.ea:14, Drumond.Rendle.ea:12,
Rodolphe.Antoine.ea:12, Yao.McCallum.ea:13}, graphical
model~\cite{Jiang.Dou.ea:12}, and graph feature based
method~\cite{Lao.Cohen.ea:11, Lao.Cohen:10} are considered. 

There exists large number of works which focus on factorization based models.
The common thread among the factorization methods is that they explain the
triples via latent features of entities. ~\cite{Bro.97} presents a tensor based
model that decomposes each entity and predicate in knowledge graphs as a low
dimensional vector.  However, such a method fails to consider the symmetry
property of the tensor.  In order to solve this issue,
~\cite{Nickel.Tresp.ea:11} proposes a relational latent feature model, RESCAL,
an efficient approach which uses a tensor factorization model that takes the
inherent structure of relational data into account. By leveraging relational
domain knowledge about entity type information, ~\cite{Chang.Yang.ea:14}
proposes a tensor decomposition approach for relation extraction in knowledge
base which is highly efficient in terms of time complexity.  In addition,
various other latent variable models, such as neural network based
methods~\cite{Dong.Lao.ea:14,Socher.Chen.ea:13}, have been explored for link
prediction task. However, the major drawback of neural network based models is
their complexity and computational cost in model training and parameter tuning.
Many of these models require tuning large number of parameters, thus finding
the right combination of these parameters is often considered more of an art
than science. 

Recently graphical models, such as Probabilistic Relational
Models~\cite{Friedman.Getoor.ea:99}, Relational Markov
Network~\cite{Taskar.Abbeel.ea:02}, Markov Logic
Network~\cite{Jiang.Dou.ea:12,Richardson.Domingos:06} have also been used for
link prediction in knowledge graph. For instance,
~\cite{Richardson.Domingos:06} proposes a Markov Logic Network (MLN) based
approach, which is a template language for defining potential functions on
knowledge graph by logical formula.  Despite its utility for modeling knowledge
graph, issues such as rule learning difficulty, tractability problem, and
parameter estimation pose implementation challenge for MLNs.

Graph feature based approaches assume that the existence of an edge can be
predicted by extracting features from the observed edges in the graph.  Lao and
Cohen ~\cite{Lao.Cohen.ea:11, Lao.Cohen:10} propose Path Ranking Algorithm
(PRA) to perform random walk on the graph and compute the probability of each
path.  The main idea of PRA is to use these path probabilities as supervised
features for each entity pair, and use any favorable classification model, such
as logistic regression and SVM, to predict the probability of missing edge
between an entity pair in a knowledge graph. 

It has been demonstrated~\cite{Bordes.Gabrilovich:14} that no single approach
emerges as a clear winner. Instead, the merits of factorization models and
graph feature models are often complementary with each other. Thus combining
the advantages of different approaches for learning knowledge graph is a
promising option.  For instance,~\cite{Nickel.Jiang.ea:14} proposes to use
additive model, which is a linear combination between RESCAL and PRA.  The
combination results in not only decrease the training time but also increase the
accuracy. ~\cite{Jiang.Huang.ea:12} combines a latent feature model with an
additive term to learn from latent and neighborhood-based information on
multi-relational data. ~\cite{Dong.Lao.ea:14} fuses the outputs of PRA and
neural network model as features for training a binary classifier.   Our work
strongly aligns with this combination approach. In this work, we build matrix
factorization based techniques that have been proved successful for recommender systems and plan to
incorporate graph based features in future work.

\section{Background and Problem Statement}

\begin{definition}{We define the knowledge graph as a collection of triple facts $G = (S, P, O)$, where $s \in S$ and $o \in O$ are the set of subject and object entities and $p \in P$ is the set of predicates or relations between them.  $G(s, p, o) = 1$ if there is a direct link of type $p$ from $s$ to $o$, and $G(s, p, o) = 0$ otherwise.}\end{definition}

Each triple fact in knowledge graph is a statement interpreted as ``A relationship p holds between entities 
$s$ and $o$". For instance, the statement ``Kobe Bryant is a player of LA Lakers" can be expressed by the following triple fact
(``Kobe Bryant", ``playsFor", ``LA Lakers"). 

\begin{definition}{For each relation $p \in P$, we define $G_{p}(S_{p}, O_{p})$ as a bipartite subgraph of $G$, where the corresponding 
set of entities $s_{p} \in S_{p}$, $o_{p} \in O_{p}$ are connected by relation $p$, namely $G_{p}(s_{p}, o_{p}) = 1$.}\end{definition}

\noindent\textbf{Problem Statement:} For every predicate $p \in P$ and given an entity pair $(s, o)$ in $G_p$, our goal is to learn a link recommendation model $M_p$ 
such that $x_{s,o} = M_p(s, o)$ is a real-valued score.


Due to the fact that the produced real-valued score is not normalized, we compute the probability $Pr(y_{s,o}^{p} = 1)$, where $y_{s, o}^{p}$
is a binary random variable that is true iff $G_{p}(s, o) = 1$. We estimate this probability $Pr$ using the logistic function as follows:
\begin{equation}
Pr(y_{s,o}^{p} = 1) = \frac{1}{1 + exp(-x_{s,o})}
\end{equation}

Thus we interpret $Pr(y_{s,o}^{p} = 1)$ as the probability that a vertex (or subject) $s$ in the knowledge graph $G$ is in a relationship
of given type $p$ with another vertex (or the object) $o$.

\section{Methods}

In this section, we describe our model, namely Latent Feature Embedding Model
with Bayesian Personalized Ranking (BPR) based optimization technique that we
propose for the task of link prediction in a knowledge graph. In our link
prediction setting, for a given predicate $p$, we first construct its bipartite
subgraph $G_p(S_p, O_p)$. Then we learn the optimal low dimensional embeddings for its
corresponding subject and object entities $s_p \in S_p$, $o_p \in O_p$
by maximizing a ranking based distance function. The learning process
relies on Stochastic Gradient Descent (SGD). The SGD based optimization
technique iteratively updates the low dimensional representation of $s_p$ and
$o_p$ until convergence. Then the learned model is used for ranking the
unobserved triple facts in descending order such that triple facts with higher score
values have a higher probability of being correct.

\subsection{Latent Feature Based Embedding Model}~\\
For each predicate $p$, the model maps both its corresponding subject and object entites $s_{p}$ and $o_{p}$
into low-dimensional continuous vector spaces, say $U_{s}^{p} \in {\rm I\!R}^{1 \times K}$ and $V_{o}^{p} \in {\rm I\!R}^{1 \times K}$ respectively. We measure the compatibility between 
subject $s_{p}$ and object $o_{p}$ as dot product of its corresponding latent vectors which is given as below:

\begin{equation}
x_{s_{p},o_{p}} = (U_{s}^{p})(V_{o}^{p})^{T} + b_{o}^{p}
\label{eq:latent}
\end{equation}

where $U^{p} \in {\rm I\!R}^{\mid S \mid \times K}$, $V^{p} \in {\rm
I\!R}^{\mid O \mid \times K}$, and $b^{p} \in {\rm I\!R}^{\mid O \mid \times
1}$.  $\left| S \right|$ and $\left| O \right|$ denote the size of subject and
object associated with predicate $p$ respectively.  $K$ is the number of latent
dimensions and $b_{o}^{p} \in {\rm I\!R}$ is a bias term associated with object
$o$. Given predicate $p$, the higher the score of $x_{s_{p},o_{p}}$, the more
similar the entities $s_{p}$ and $o_{p}$ in the embedded low dimensional space,
and the higher the confidence to include this triple fact into knowledge
base.

\subsection{Bayesian Personalized Ranking}~\\
In collaborative filtering, positive-only data is known as implicit feedback/binary feedback. For example, in the eCommerce platform,
some users only buy but do not rate items. Motivated by~\cite{Rendle.Gantner.ea:09}, we employ Bayesian Personalized Ranking (BPR) based
approach for model learning. Specifically, in recommender system domain, given user-item matrix, BPR based approach assigns the preference of user for purchased 
item with higher score than un-purchased item. Likewise, under this context, we assign observed triple facts higher score
than unobserved triple facts in knowledge base. We assume that unobserved facts are not necessarily negative, rather they are ``less preferable" than the observed ones. 

For our task, in each predicate $p$, we denote the observed subject/object entity pair as $(s_p, o^{+}_p)$ and unobserved one
as $(s_p, o^{-}_p)$. The observed facts in our case are the existing link between $s_p$ and $o_p$ given $G_p$ and unobserved ones are the 
missing link between them. Given this fact, BPR maximizes the following ranking based distance function:

\begin{equation}
\scalebox{0.88} {$BPR = \underset{\Theta_{p}}{\text{max}} \\ \sum_{(s_p,o^{+}_p,o^{-}_p) \in D_{p}} \ln\sigma(x_{s_p,o^{+}_p} - x_{s_p,o^{-}_p}) - \lambda_{\Theta_{p}}\mid\mid\Theta_{p}\mid\mid^{2}$}
\label{eq:objbpr}
\end{equation}

where $D_{p}$ is a set of samples generated from the training data for predicate $p$, $G_{p}(s_p, o^{+}_p) = 1$ and $G_{p}(s_p, o^{-}_p) = 0$. 
And $x_{s_p,o^{+}_p}$ and $x_{s_p,o^{-}_p}$ are the predicted scores of 
subject $s_p$ on objects $o^{+}_p$ and $o^{-}_p$ respectively. We use the proposed
latent feature based embedding model shown in Equation~\ref{eq:latent} to compute $x_{s_p, o^{+}_p}$ and $x_{s_p, o^{-}_p}$ respectively. 
The last term in Equation~\ref{eq:objbpr} is a $l_{2}$-norm regularization term used for 
model parameters $\Theta_{p} = \{U^{p}, V^{p}, b^{p}\}$ to avoid overfitting in the learning process. 
In addition, the logistic function $\sigma(.)$ in Equation~\ref{eq:objbpr} is defined as $\sigma(x) = \frac{1}{1+e^{-x}}$.


Notice that the Equation~\ref{eq:objbpr} is differentiable, thus we employ the widely used SGD to maximize the objective.
In particular, at each iteration, for given predicate $p$, we sample one observed entity pair $(s_p, o^{+}_p)$ and one unobserved one
$(s_p, o^{-}_p)$ using uniform sampling technique. Then we iteratively update the model parameters $\Theta_{p}$ based on the sampled pairs. 
Specifically, for each training instance, we compute the derivative and update the corresponding parameters $\Theta_{p}$ by walking along the ascending gradient direction. 

For each predicate $p$, given a training triple $(s_p, o^{+}_p, o^{-}_p)$, the gradient of BPR objective in Equation~\ref{eq:objbpr} with respect to
$U_{s}^{p}$, $V_{o^{+}}^{p}$, $V_{o^{-}}^{p}$, $b_{o^{+}}^{p}$, $b_{o^{-}}^{p}$ can be computed as follows:

\begin{eqnarray}
\frac{\partial BPR}{\partial U_{s}^{p}} &=& \frac{\partial \ln\sigma(x_{s_p,o^{+}_p} - x_{s_p,o^{-}_p})}{\partial U_{s}^{p}} - 2\lambda^{p}_{s}U_{s}^{p} \nonumber \\ 
&=& \scalebox{0.98} {$\frac{\partial \ln\sigma(x_{s_p,o^{+}_p} - x_{s_p, o^{-}_p})}{\partial \sigma(x_{s_p,o^{+}_p} - x_{s_p, o^{-}_p})} \times \frac{\partial\sigma(x_{s_p,o^{+}_p} - x_{s_p, o^{-}_p})}{\partial(x_{s_p,o^{+}_p} - x_{s_p, o^{-}_p})}$} \nonumber\\
&& \scalebox{0.98} {$\times \frac{\partial(x_{s_p,o^{+}_p} - x_{s_p, o^{-}_p})}{\partial U_{s}^{p}} - 2\lambda^{p}_{s}U_{s}^{p}$} \nonumber \\
&=& \frac{1}{\sigma(x_{s_p,o^{+}_p} - x_{s_p,o^{-}_p})} \times \sigma(x_{s_p,o^{+}_p} - x_{s_p,o^{-}_p}) \nonumber\\
&&\scalebox{0.90} {$\big(1-\sigma(x_{s_p,o^{+}_p} - x_{s_p,o^{-}_p})\big)\times(V_{o^{+}}^{p} - V_{o^{-}}^{p}) - 2\lambda^{p}_{s}U_{s}^{p}$} \nonumber \\
&=&\scalebox{0.90} {$\big(1 - \sigma(x_{s_p,o^{+}_p} - x_{s_p,o^{-}_p})\big)(V_{o^{+}}^{p} - V_{o^{-}}^{p}) - 2\lambda^{p}_{s}U_{s}^{p}$} \nonumber \\
\label{eq:update1}
\end{eqnarray}

We obtain the following using similar chain rule derivation.

\begin{equation}
\frac{\partial BPR}{\partial V_{o^{+}}^{p}} = \big(1 - \sigma(x_{s_p,o^{+}_p} - x_{s_p,o^{-}_p})\big)\times U_{s}^{p} - 2\lambda^{p}_{o^{+}}V_{o^{+}}^{p}
\end{equation}

\begin{equation}
\frac{\partial BPR}{\partial V_{o^{-}}^{p}} = \big(1 - \sigma(x_{s_p,o^{+}_p} - x_{s_p,o^{-}_p})\big)\times (-U_{s}^{p}) - 2\lambda^{p}_{o^{-}}V_{o^{-}}^{p}
\end{equation}

\begin{equation}
\frac{\partial BPR}{\partial b_{o^{+}}^{p}} = \big(1 - \sigma(x_{s_p,o^{+}_p} - x_{s_p,o^{-}_p})\big)\times 1 - 2\lambda^{p}_{o^{+}}b_{o^{+}}^{p}
\end{equation}

\begin{equation}
\frac{\partial BPR}{\partial b_{o^{-}}^{p}} = \big(1 - \sigma(x_{s_p,o^{+}_p} - x_{s_p,o^{-}_p})\big)\times (-1) - 2\lambda^{p}_{o^{-}}b_{o^{-}}^{p}
\end{equation}

Next, the parameters are updated as follows:

\begin{equation}
U_{s}^{p} = U_{s}^{p} + \alpha\times\frac{\partial BPR}{\partial U_{s}^{p}} 
\label{eq:sgd1}
\end{equation}

\begin{equation}
V_{o^{+}}^{p} = V_{o^{+}}^{p} + \alpha\times\frac{\partial BPR}{\partial V_{o^{+}}^{p}} 
\label{eq:sgd2}
\end{equation}

\begin{equation}
V_{o^{-}}^{p} = V_{o^{-}}^{p} + \alpha\times\frac{\partial BPR}{\partial V_{o^{-}}^{p}} 
\label{eq:sgd3}
\end{equation}

\begin{equation}
b_{o^{+}}^{p} = b_{o^{+}}^{p} + \alpha\times\frac{\partial BPR}{\partial b_{o^{+}}^{p}} 
\label{eq:sgd4}
\end{equation}

\begin{equation}
b_{o^{-}}^{p} = b_{o^{-}}^{p} + \alpha\times\frac{\partial BPR}{\partial b_{o^{-}}^{p}} 
\label{eq:sgd5}
\end{equation}

where $\alpha$ is the learning rate.

\begin{algorithm}
\renewcommand{\algorithmicrequire}{\textbf{Input:}}
\renewcommand{\algorithmicensure}{\textbf{Output:}}
\caption{Bayesian Personalized Ranking Based Latent Feature Embedding Model}
\label{alg:1}
\begin{algorithmic}[1]
\REQUIRE latent dimension $K$, $G$, target predicate $p$
\ENSURE $U^{p}$, $V^{p}$, $b^{p}$
\STATE Given target predicate $p$ and entire knowledge graph $G$, construct its bipartite subgraph, $G_{p}$ 
\STATE $m$ = number of subject entities in $G_{p}$
\STATE $n$ = number of object entities in $G_{p}$ 
\STATE Generate a set of training samples $D_{p} = \{(s_p, o^{+}_{p}, o^{-}_{p})\}$ using uniform sampling technique
\STATE Initialize $U^{p}$ as size $m \times K$ matrix with $0$ mean and standard deviation $0.1$
\STATE Initialize $V^{p}$ as size $n \times K$ matrix with $0$ mean and stardard deviation $0.1$
\STATE Initialize $b^{p}$ as size $n \times 1$ column vector with $0$ mean and stardard deviation $0.1$
\FORALL{$(s_p, o^{+}_{p}, o^{-}_{p}) \in D_{p}$}
  \STATE Update $U_{s}^{p}$ based on Equation~\ref{eq:sgd1}
  \STATE Update $V_{o^{+}}^{p}$ based on Equation~\ref{eq:sgd2}
  \STATE Update $V_{o^{-}}^{p}$ based on Equation~\ref{eq:sgd3}
  \STATE Update $b_{o^{+}}^{p}$ based on Equation~\ref{eq:sgd4}
  \STATE Update $b_{o^{-}}^{p}$ based on Equation~\ref{eq:sgd5}
\ENDFOR
\STATE \textbf{return} $U^{p}$, $V^{p}$, $b^{p}$
\end{algorithmic}  
\end{algorithm}

\subsection{Pseudo-code and Complexity Analysis}~\\
The pseudo-code of our proposed link prediction model is described in Algorithm~\ref{alg:1}.
It takes the knowledge graph $G$ and a specific target predicate $p$ as input and generates the low
dimensional latent matrices $U^{p}$, $V^{p}$, $b^{p}$ as output. Line 1 constucts the bipartite subgraph
of predicate $p$, $G_{p}$ given entire knowledge graph $G$. Line 2-3 compute the number of subject and object entities
as $m$ and $n$ in resultant bipartite subgraph $G_{p}$ respectively. Line 4 generates a collection of triple samples using 
uniform sampling technique. Line 5-7 initialize the matrices $U^{p}$, $V^{p}$, $b^{p}$ using 
Gaussian distribution with $0$ mean and $0.1$ standard deviation, assuming all the entries in $U^{p}$, $V^{p}$ and $b^{p}$ are independent.
Line 8-14 update corresponding rows of matrices $U^{p}$, $V^{p}$, $b^{p}$ based on the sampled instance $(s_p, o^{+}_{p}, o^{-}_{p})$ in each iteration. 
As the sample generation step in line 4 is prior to the model parameter learning, thus the convergence criteria of Algorithm~\ref{alg:1} is to iterate
over all the sampled triples in $D_{p}$.  

Given the constructed $G_{p}$ as input, the time complexity of the update rules shown in 
Equations~\ref{eq:sgd1}~\ref{eq:sgd2}~\ref{eq:sgd3}~\ref{eq:sgd4}~\ref{eq:sgd5} is $\mathcal{O}(cK)$, where $K$ is the number of latent features. 
The total computational complexity of Algorithm~\ref{alg:1} is then $\mathcal{O}(\left|D_{p}\right|\cdot cK)$, 
where $\left| D_{p}\right|$ is the total size of pre-sampled triples shown in line 4 of Algorithm~\ref{alg:1}.

\section{Experiments and Results}
This section presents our experimental analysis of the Algorithm~\ref{alg:1} for thirteen unique predicates in the well known YAGO2 knowledge graph~\cite{Hoffart.Suchanek.ea:11}.
We construct a model for each predicate and describe our evaluation strategies, 
including performance metrics and selection of state-of-the-art methods for benchmarking in section 5.1. We aim to answer two questions through our experiments:

\begin{figure*}
\centering
\subfigure[HR Comparison among different link recommendation methods]{\label{fig:a}\includegraphics[width=56mm]{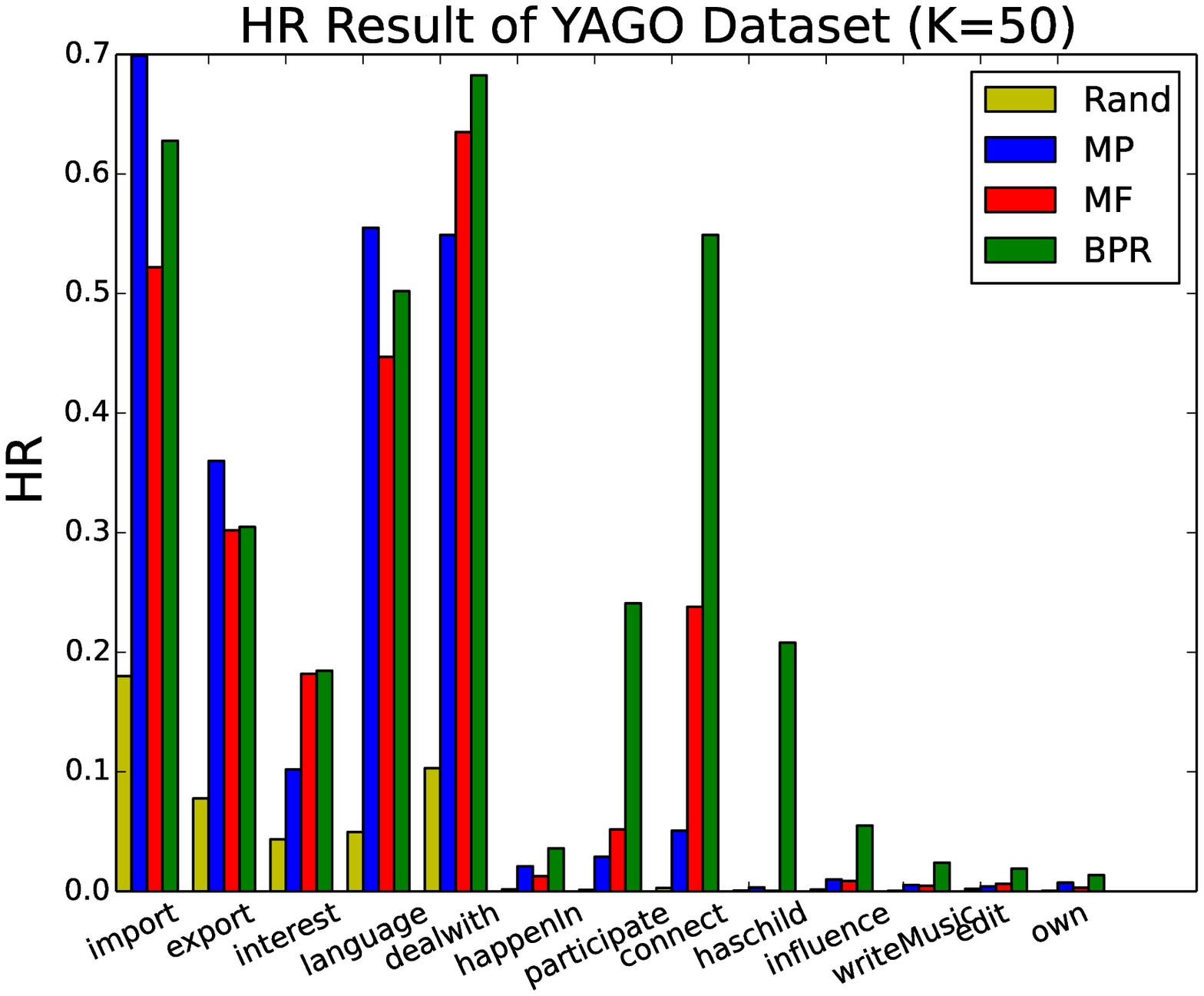}}
\subfigure[ARHR Comparison among different link recommendation methods]{\label{fig:b}\includegraphics[width=56mm]{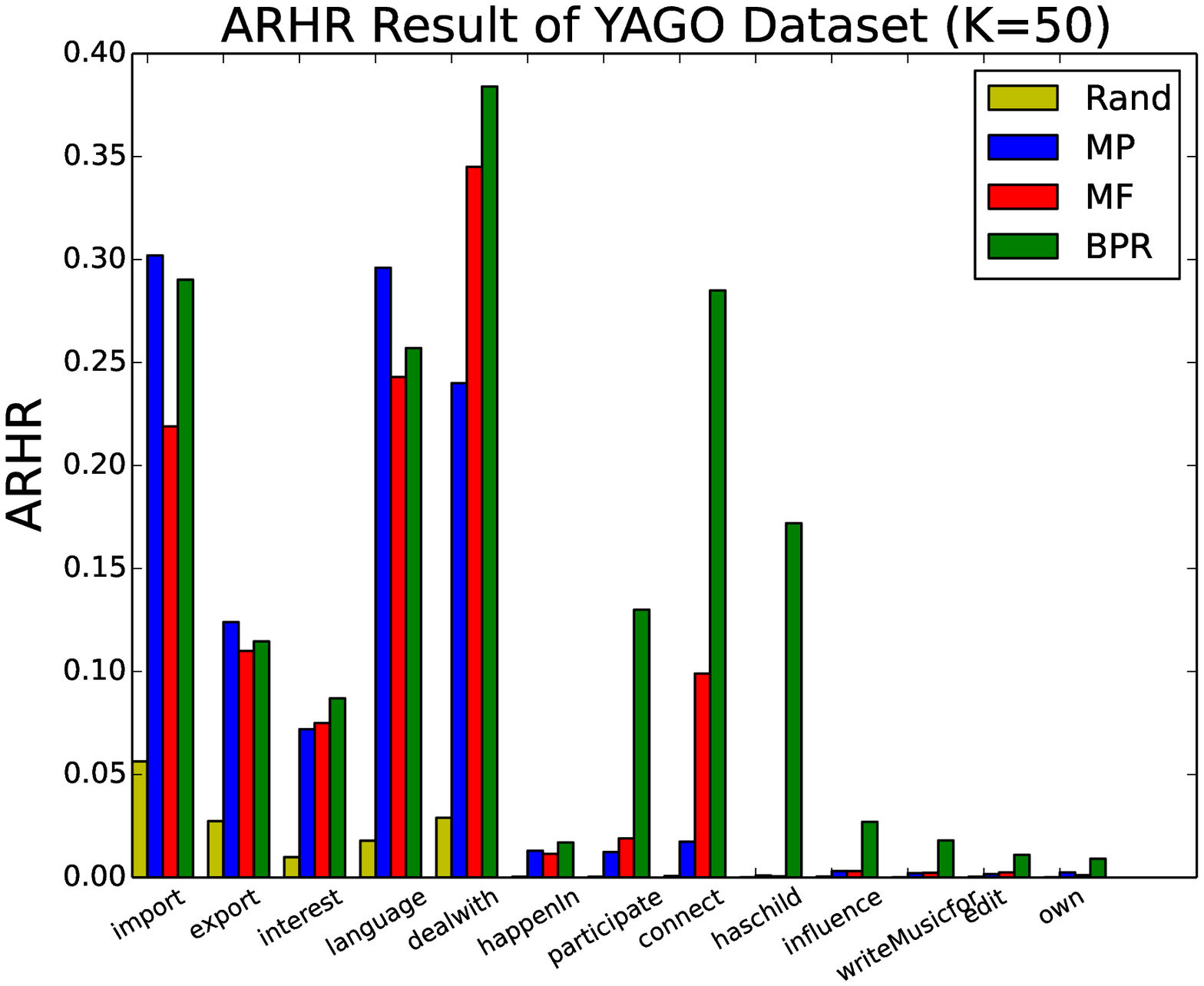}}        
\subfigure[AUC Comparison among different link recommendation methods]{\label{fig:c}\includegraphics[width=56mm]{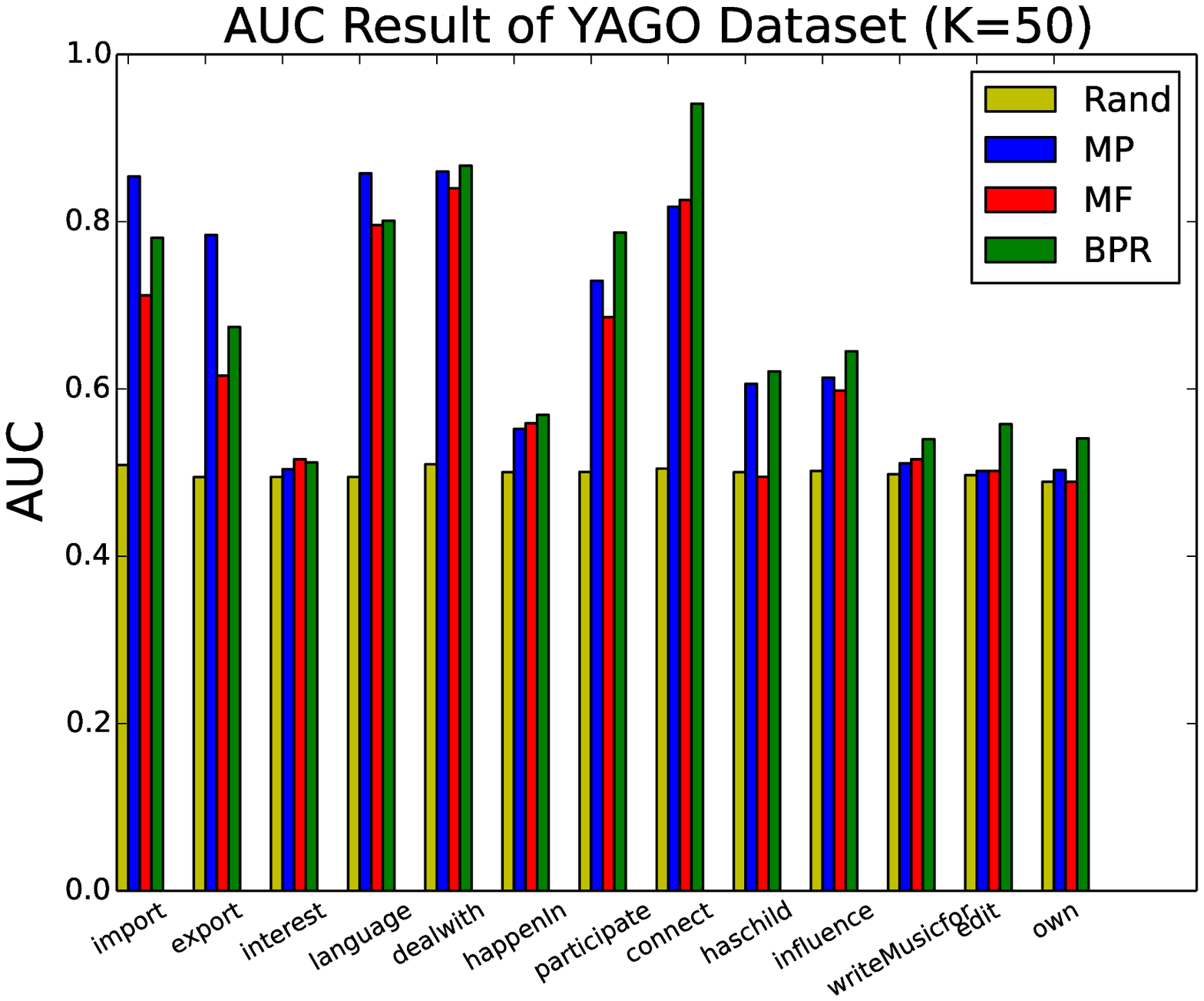}}
\caption{Link Recommendation Comparison on YAGO2 Relations}\label{fig:yagoprediction}
\end{figure*}

\begin{enumerate}

\item How does our approach compare with related work for link recommendation in knowledge graph?

\item For a predicate $p$, can we reason about the link prediction model performance $M_p$ in terms of the structural metrics of the bipartite graph $G_p$?

\end{enumerate}

\begin{table}[h!]
\centering
\scalebox{0.75}{
\begin{tabular}{*{4}{c}}
\toprule
Relation & \# Subjects & \# Objects & \# of Facts in YAGO2 \\ 
\midrule
Import & 142 & 62 & 391 \\
Export & 140 & 176 & 579 \\
isInterestedIn & 358 & 213 & 464 \\
hasOfficialLanguage & 583 & 214 & 964 \\
dealsWith & 131 & 124 & 945 \\
happenedIn & 7121 & 5526 & 12500 \\
participatedIn & 2330 & 7043 & 16809 \\
isConnectedTo & 2835 & 4391 & 33581 \\
hasChild & 10758 & 12800 & 17320 \\
influence & 8056 & 9153 & 25819 \\
wroteMusicFor & 5109 & 21487 & 24271 \\
edited & 549 & 5673 & 5946 \\
owns & 8330 & 24422 & 26536 \\
\bottomrule
\end{tabular}} 
\caption{Statistics of Various Relations in YAGO2 Dataset}
\vspace{-0.10in}
\label{tab:yagosta}
\end{table}

Table~\ref{tab:yagosta} shows the statistic of various YAGO2 relations used in our experiments. \# Subjects and \# Objects represent the number of subject and object entities associated with its corresponding predicate. The last column shown in Table~\ref{tab:yagosta} shows the number of facts for each relation in YAGO2. We run all the experiments on a 2.1 GHz Machine with 4GB memory running Linux operating system. The algorithms are implemented in Python language along with NumPy and SciPy libraries for linear algebra operations. The software is available online for download~\footnote{\url{https://sites.google.com/site/baichuanzhangpurdue}}.

\begin{figure*}
\centering
\subfigure[Graph Density and HR]{\label{fig:a}\includegraphics[width=56mm]{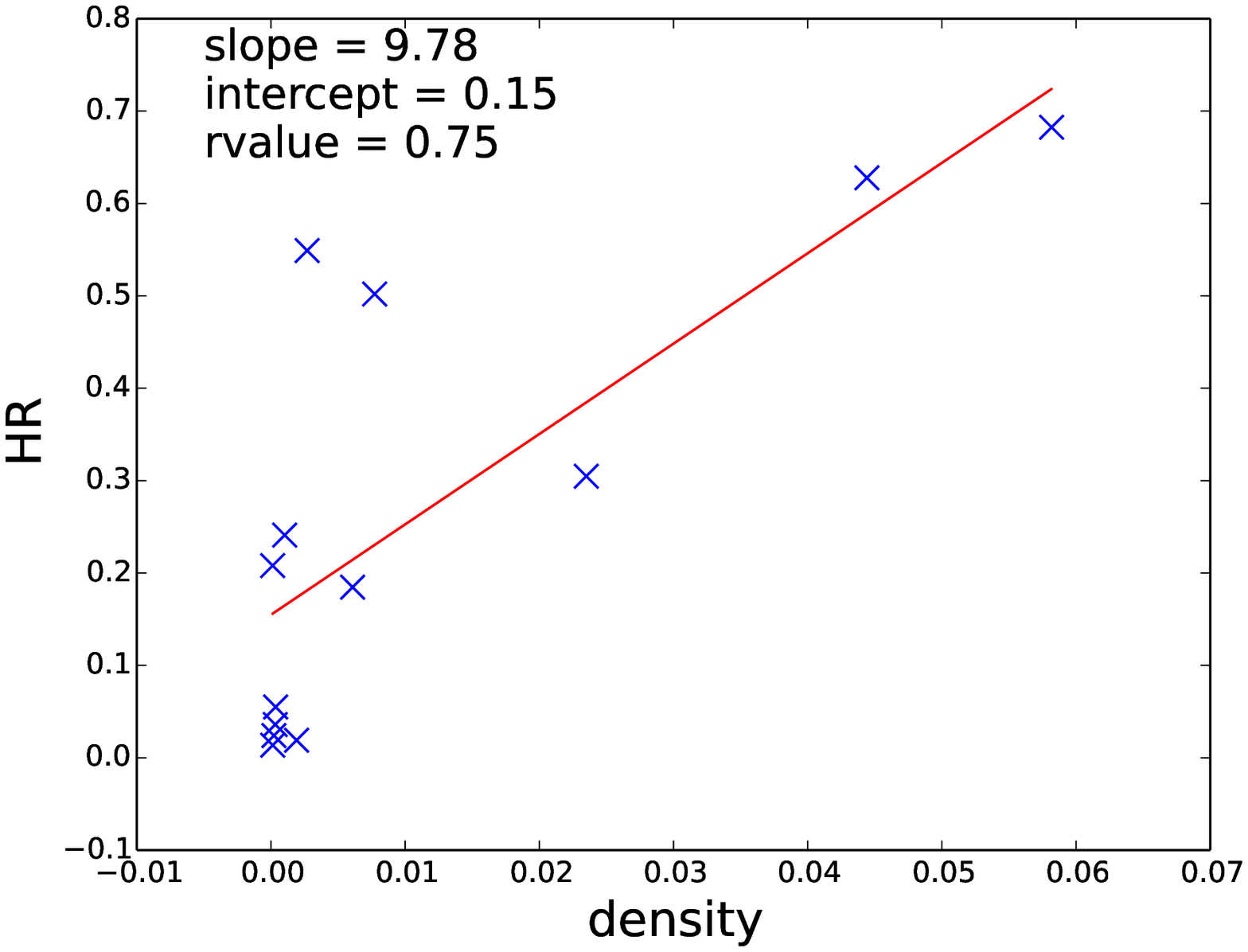}}
\subfigure[Graph Density and ARHR]{\label{fig:b}\includegraphics[width=56mm]{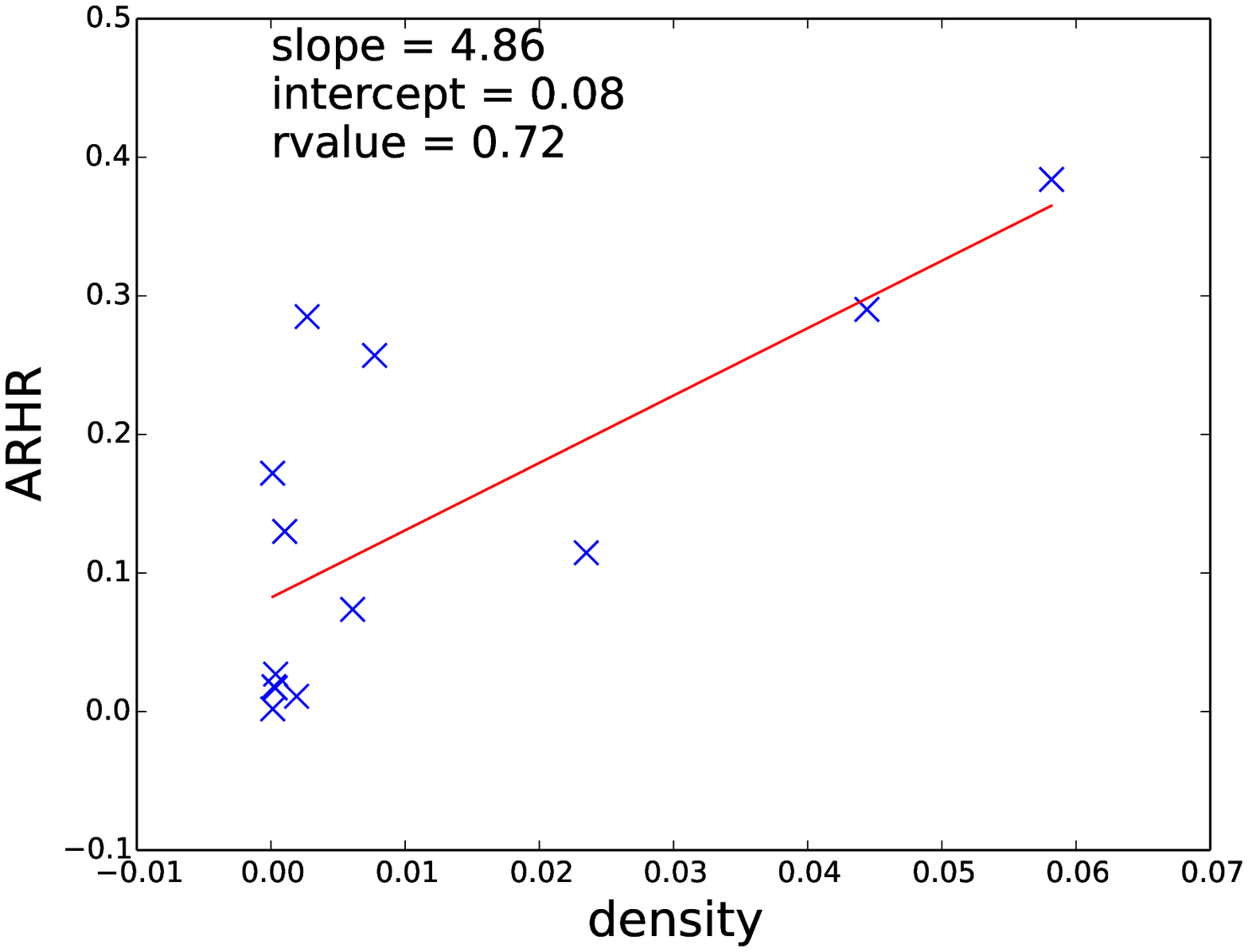}}        
\subfigure[Graph Density and AUC]{\label{fig:c}\includegraphics[width=56mm]{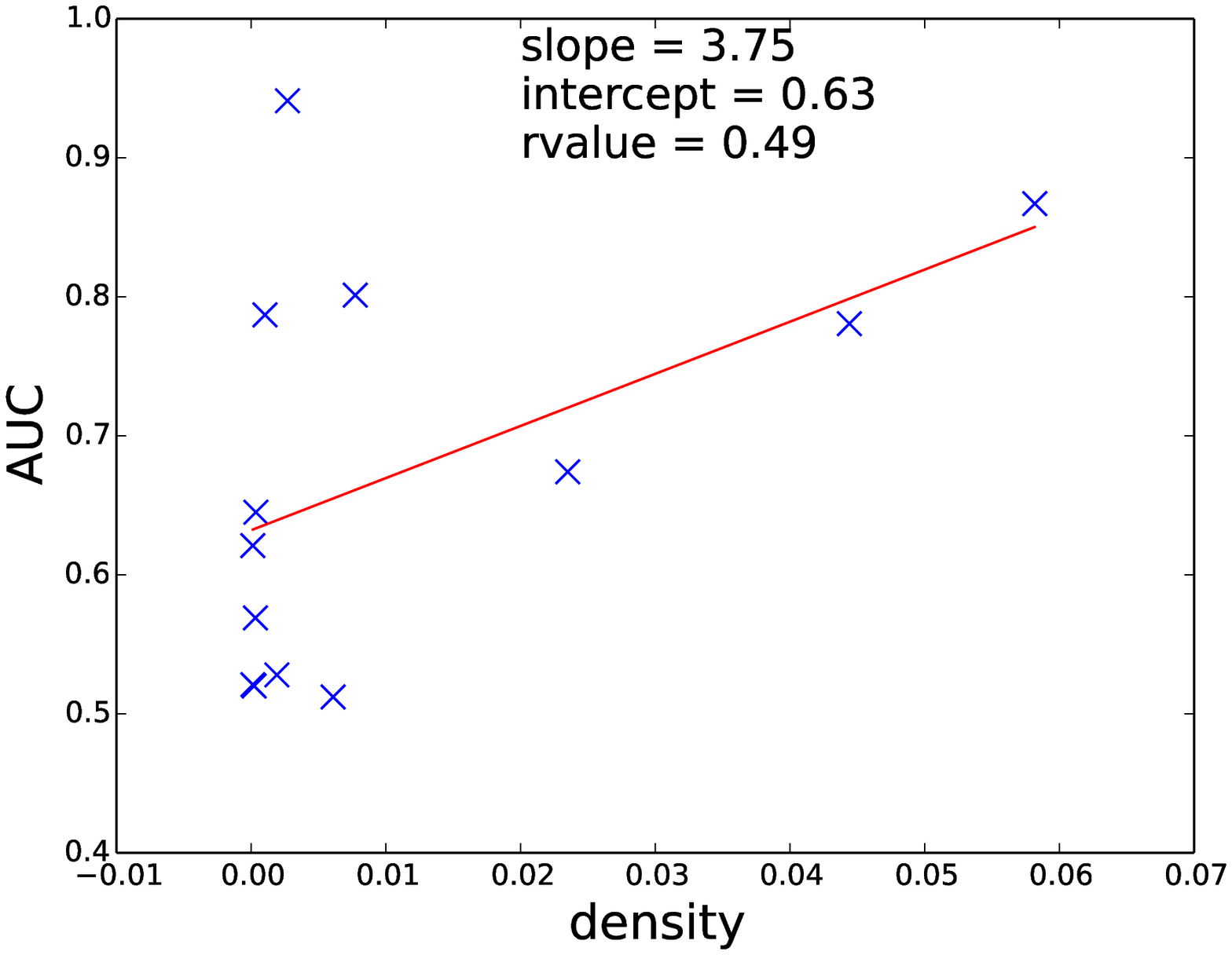}}
\subfigure[Graph Average Degree and HR]{\label{fig:a}\includegraphics[width=56mm]{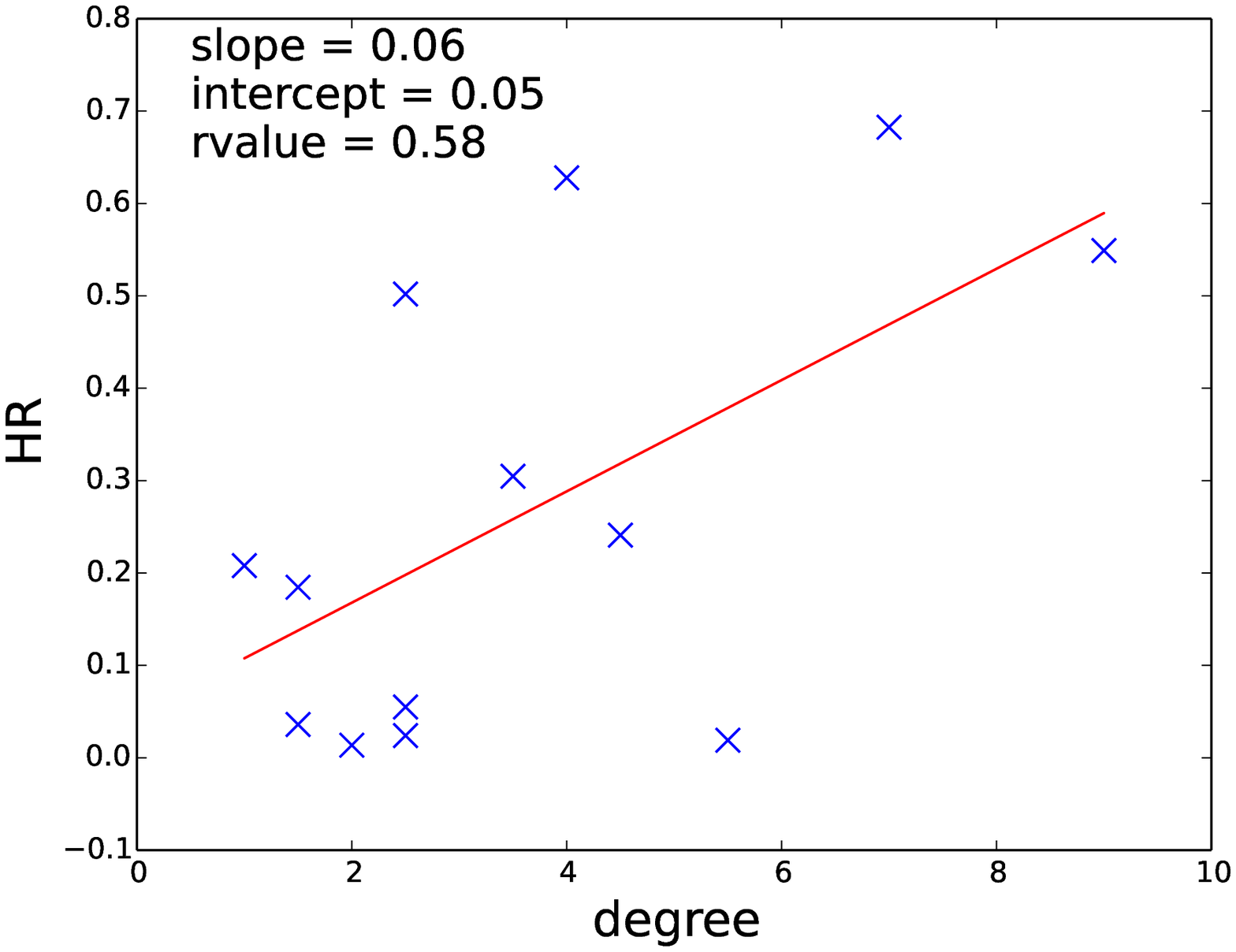}}
\subfigure[Graph Average Degree and ARHR]{\label{fig:b}\includegraphics[width=56mm]{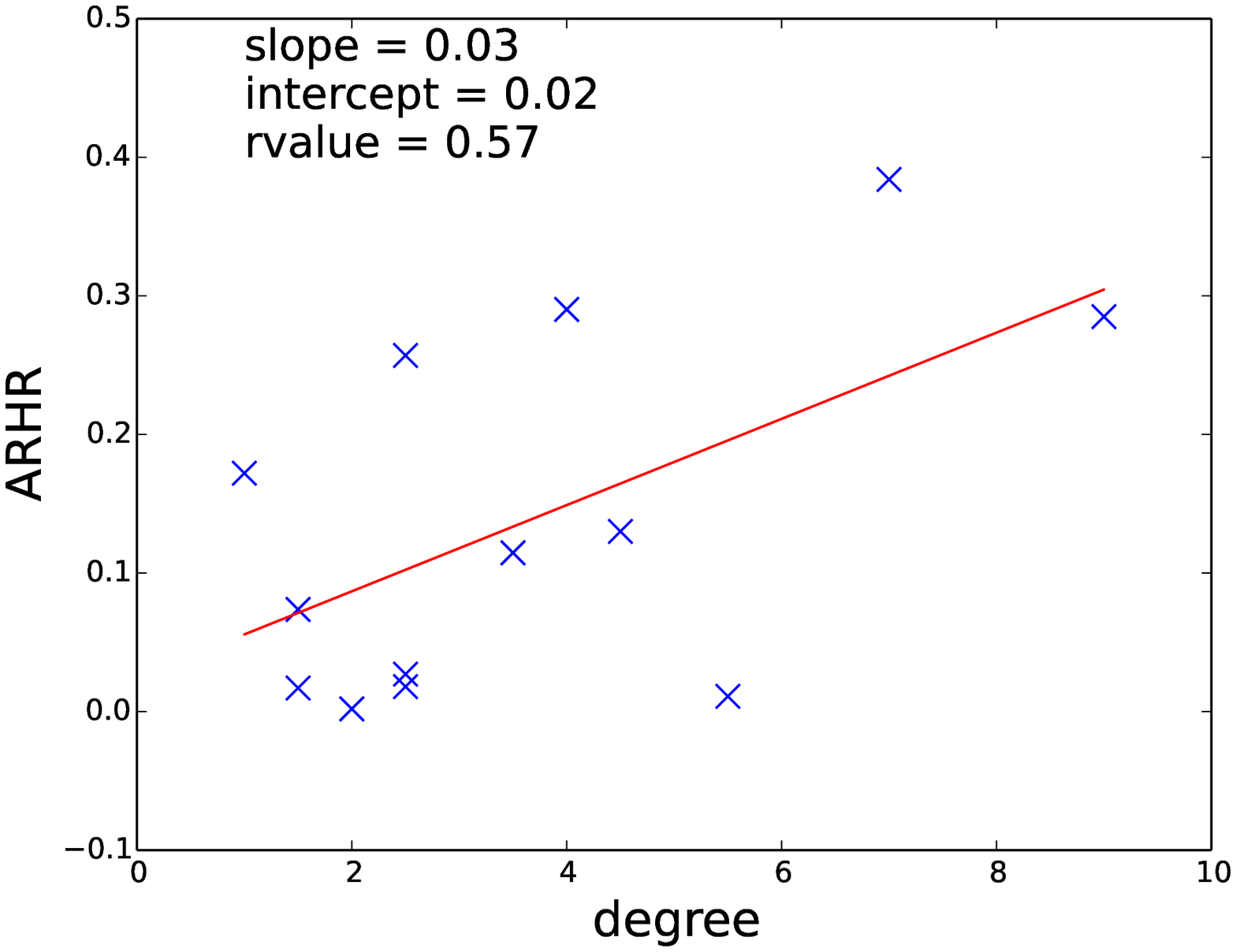}}        
\subfigure[Graph Average Degree and AUC]{\label{fig:c}\includegraphics[width=56mm]{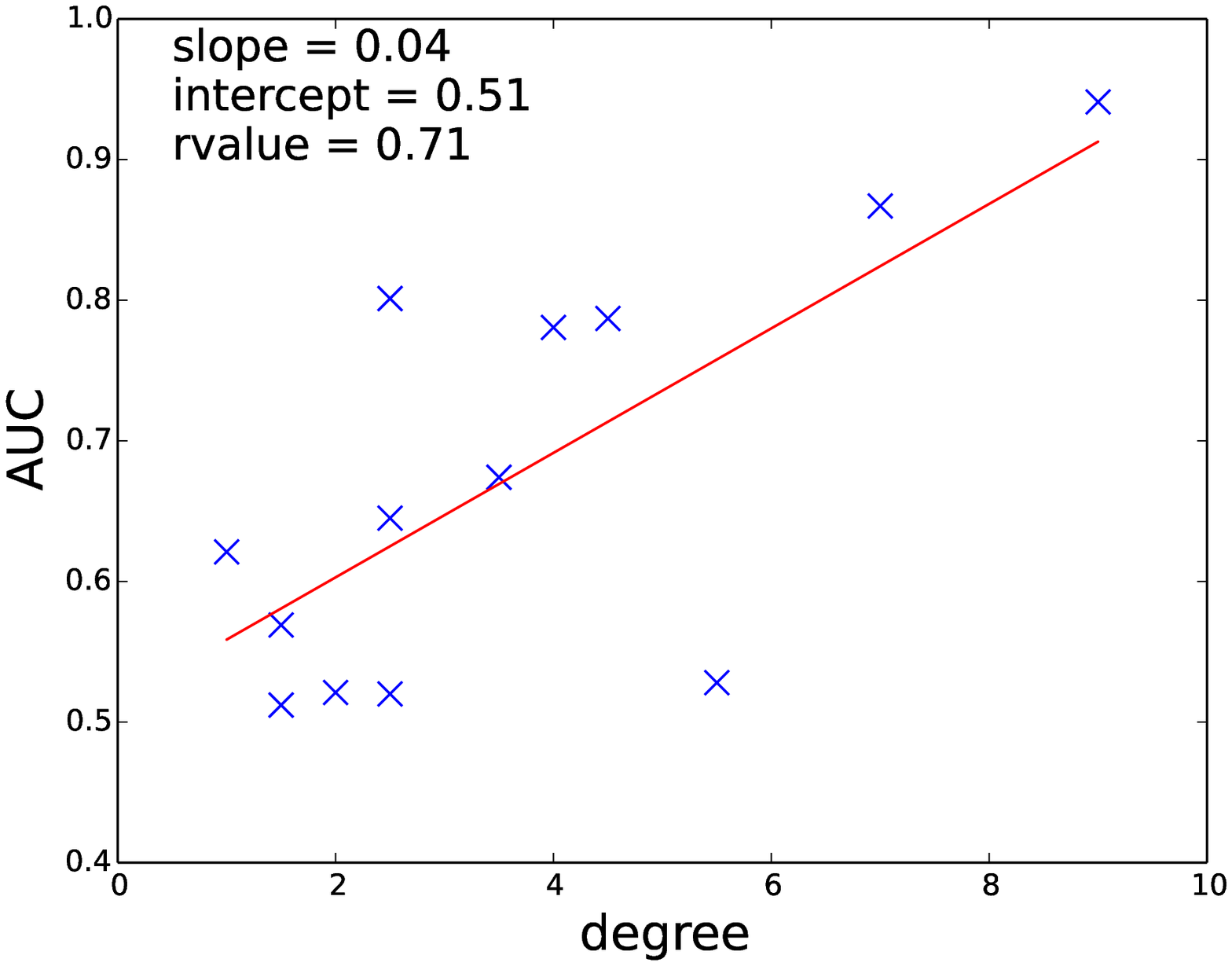}}
\subfigure[Clustering Coefficient and HR]{\label{fig:a}\includegraphics[width=56mm]{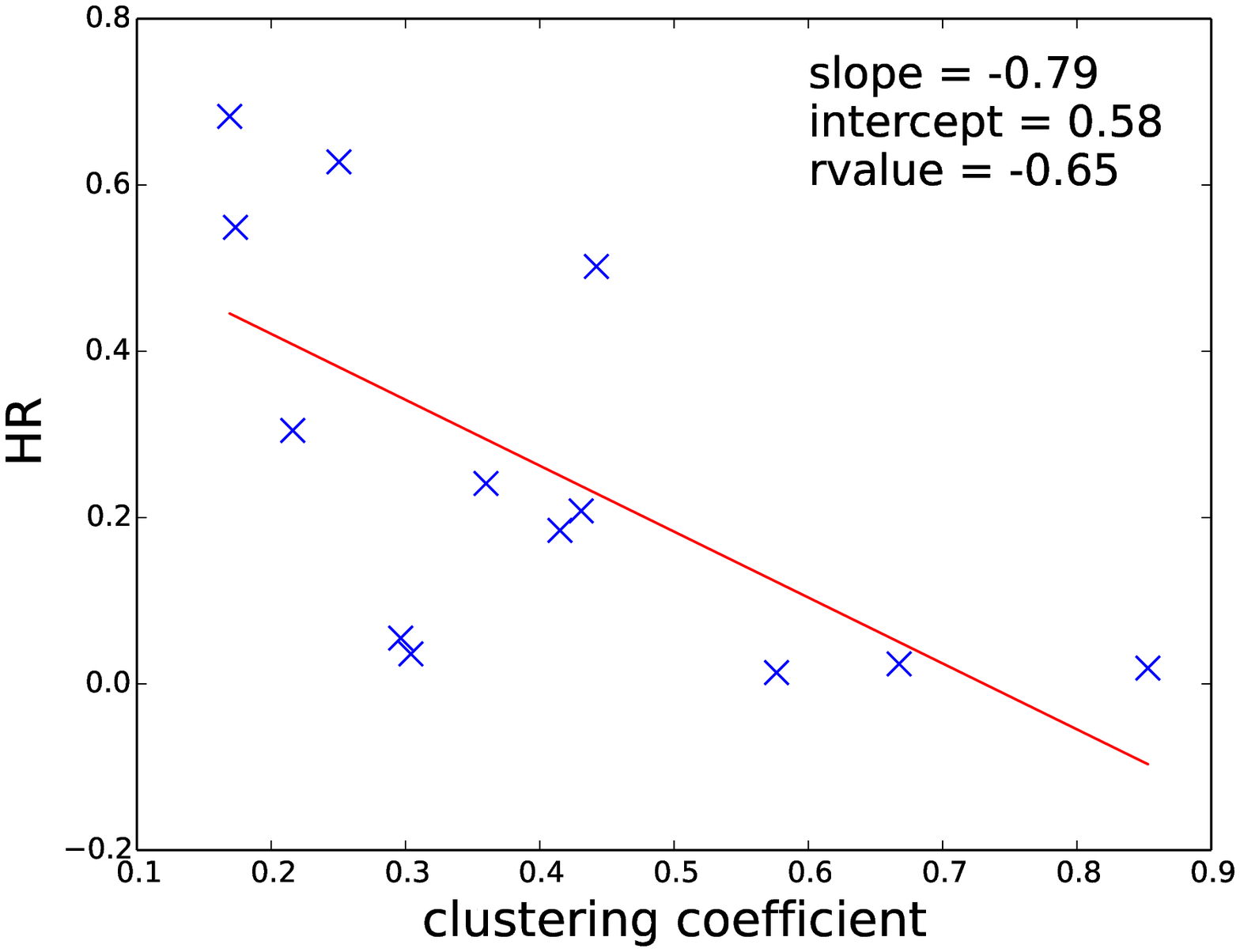}}
\subfigure[Clustering Coefficient and ARHR]{\label{fig:b}\includegraphics[width=56mm]{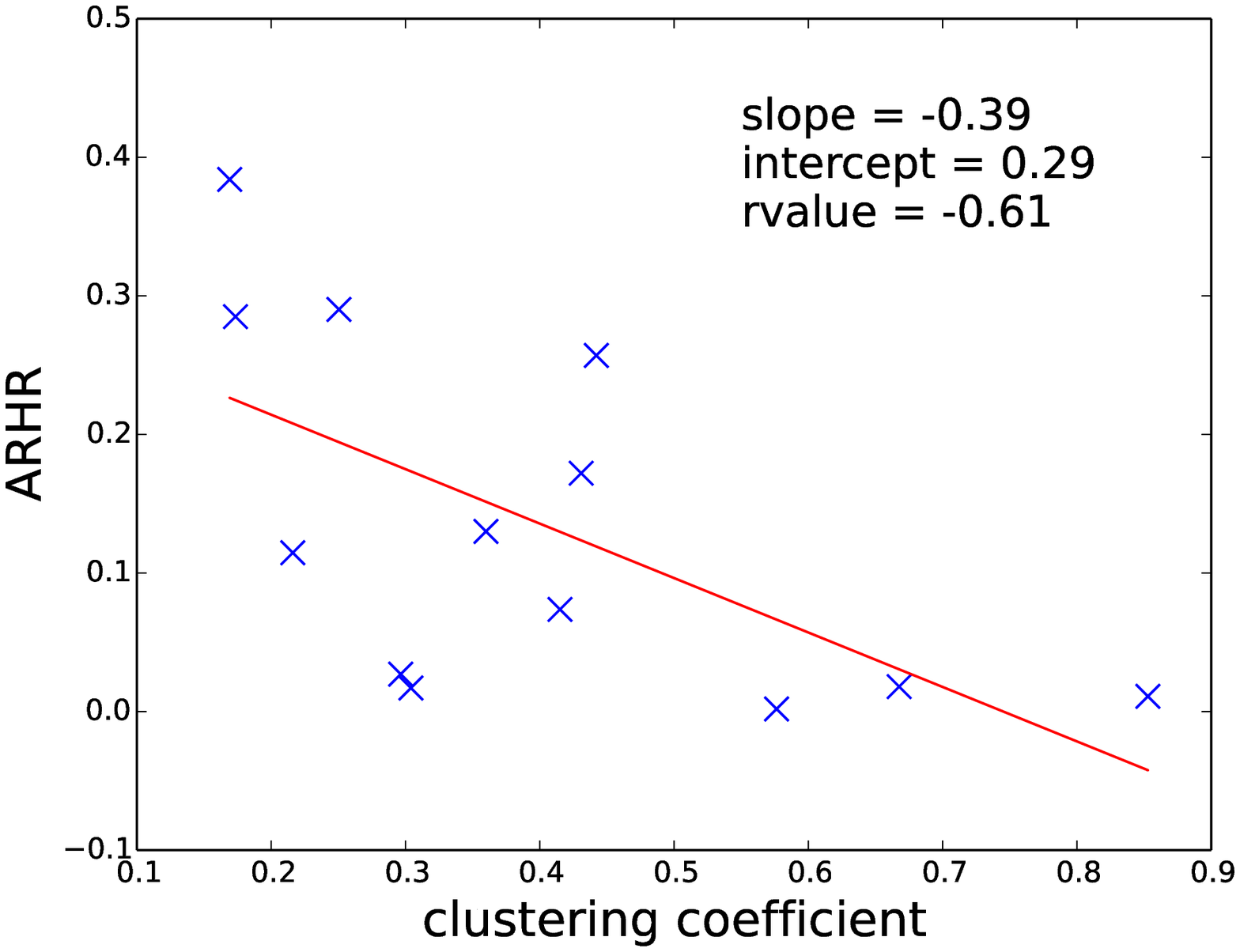}}        
\subfigure[Clustering Coefficient and AUC]{\label{fig:c}\includegraphics[width=56mm]{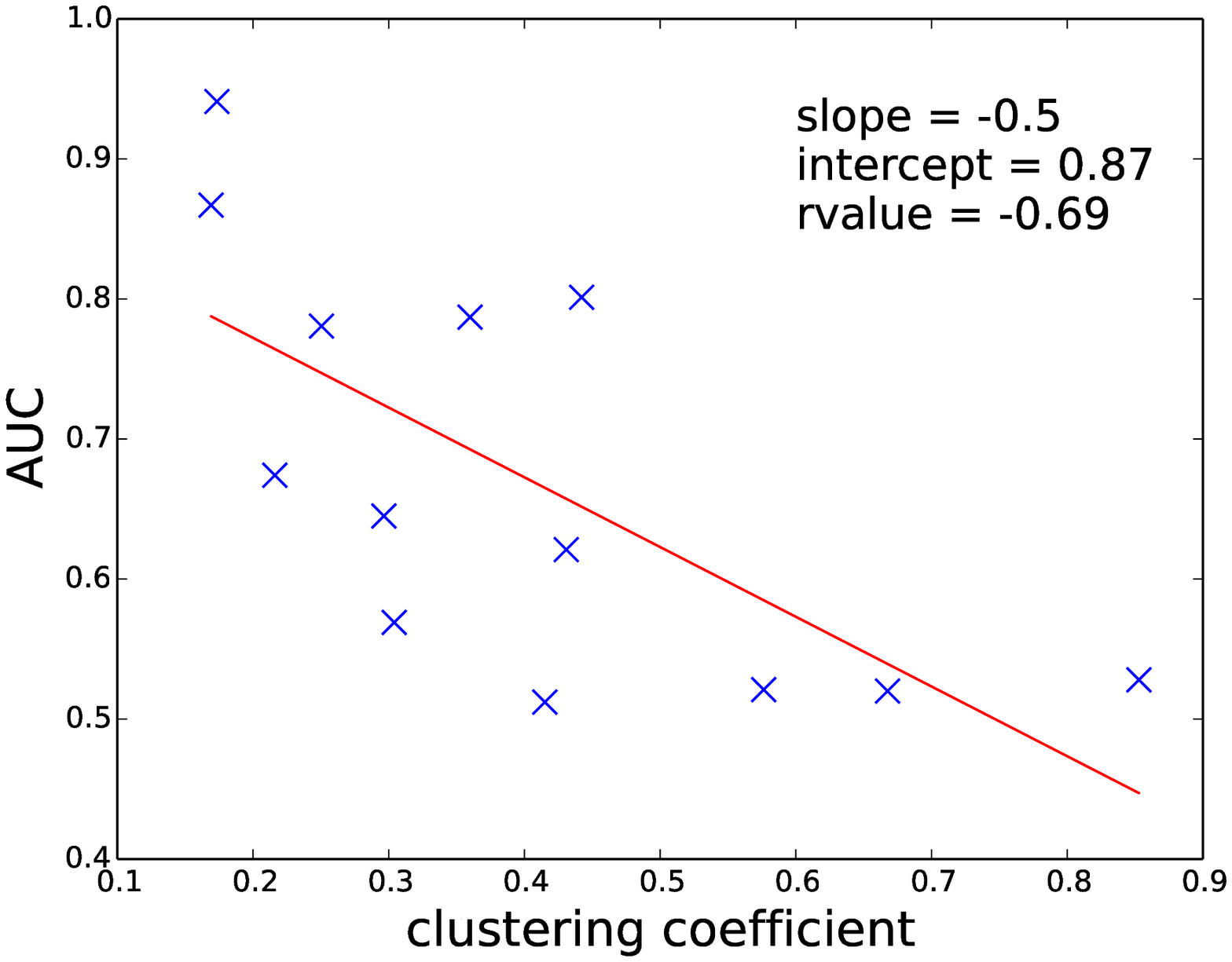}}
\caption{Quantitative Analysis Between Graph Topology and Link Recommendation Model Performance}\label{fig:topology}
\end{figure*}

\subsection{Experimental Setting}~\\
For our experiment, in order to demonstrate the performance of our proposed link prediction model, we use the YAGO2 dataset
and several evaluation metrics for all compared algorithms. Particularly, for each relation, we split the data into a training part, 
used for model training, and a test part, used for model evaluation. We apply 5-time leave one out evaluation strategy, where for
each subject, we randomly remove one fact (one subject-object pair) and place it into test set $S_{test}$ and remaining in the training
set $S_{train}$. For every subject, the training model will generate a size-N ranked list of recommended objects for 
recommendation task. The evaluation is conducted by comparing the recommendation list of each subject and the
object entity of that subject in the test set. Grid search is applied to find regularization parameters, and we set the values of parameters used in section 4.2 as 
$\lambda_s = \lambda_{o^{+}} = \lambda_{o^{-}} = 0.005$. For other model parameters, we fix learning rate
$\alpha = 0.2$, and number of latent factors $K = 50$ respectively. 
For parameter in model evaluation, we set $N = 10$. 

In order to illustrate the merit of our proposed approach, we compare our model with the following methods
for link prediction in a knowledge graph. Since the problem we solve in this paper is similar to the one-class
item recommendation~\cite{Rendle.Gantner.ea:09} in recommender system domain, we consider the following state-of-the-art one-class
recommendation methods as baseline approaches for comparison. 

\begin{enumerate}

\item \textbf{Random (Rand):} For each relation, this method randomly selects subject-object entity pair for link recommendation task.

\item \textbf{Most Popular (MP):} For each predicate in knowledge base, this method presents a non-personalized ranked object list 
based on how often object entities are connected among all subject entities. 

\item \textbf{MF:} The matrix factorization method is proposed by~\cite{Koren.Bell.ea:09}, which uses a point-wise strategy for solving 
the one-class item recommendation problem. 

\end{enumerate} 

During the model evaluation stage, we use three popular metrics, namely Hit Rate (HR), Average Reciprocal Hit-Rank (ARHR), and
Area Under Curve (AUC), to measure the link recommendation quality of our proposed approach in comparison to baseline methods.
HR is defined as follows:

\begin{equation}
HR = \frac{\#hits}{\#subjects}
\end{equation}

where $\#subjects$ is the total number of subject entities in test set, and $\#hits$ is the number of subjects
whose object entity in the test set is recommended in the size-N recommendation list. The second evaluation metric,
ARHR, considering the ranking of the recommended object for each subject entity in knowledge graph, is defined  as
below:

\begin{equation}
ARHR = \frac{1}{\#subjects}\sum_{i=1}^{\#hits}\frac{1}{p_i}
\end{equation} 

where if an object of a subject is recommended for connection in knowledge graph which we name as hit under this scenario,
$p_i$ is the position of the object in the ranked recommendation list. As we can see, ARHR is a weighted version of HR and it
captures the importance of recommended object in the recommendation list.

The last metric, AUC is defined as follows:

\begin{equation}
\scalebox{0.85} {$AUC = \frac{1}{\#subjects}\sum_{s \in subjects} \frac{1}{\mid E(s) \mid}\sum_{(o^{+},o^{-}) \in E(s)} \delta(x_{s,o^{+}} > x_{s,o^{-}})$}
\end{equation}

Where $E(s) = \{(o^{+}, o^{-}) |(s,o^{+})\in S_{test} \cap (s,o^{-}) \not \in (S_{test} \cup S_{train}) \}$, and
$\delta()$ is the indicator function.

For all of three metrics, higher values indicate better model performance. Specifically, the trivial AUC of a random predictor
is 0.5 and the best value of AUC is 1.

\subsection{YAGO2 Relation Prediction Performance}~\\



Figure~\ref{fig:yagoprediction} shows the average link prediction performance for YAGO2 relations using various methods.
Our proposed latent feature embedding approach shows overall improvement compared with other algorithms on most
of relations in YAGO2. For instance, for all the YAGO2 predicates used in the experiment, our proposed  model consistently outperforms
MF based method, which demonstrates the empirical experience that pairwise ranking based method achieves much better performance than pointwise 
regression based method given implicit feedback for link recommendation task. Compared with Popularity based recommendation method MP, our method obtains
better performance for most predicates. For example, predicates such as ``participate",``connect",``hasChild", 
and ``influence", our proposed model achieves more than 10 times better performance in terms of both HR and ARHR. However, for several predicates
such as ``import", ``export", and ``language", MP based method performs the best among all the competing methods. The good performance of MP
is owing to the semantic meaning of specific predicate. For instance, ``import" represents Country/Product relation in YAGO2, which indicates the types 
of its subject and object entities are geographic region and commodity respectively. For such a predicate, most popular
object entities such as food, cloth, fuel are linked to most of the countries, which helps MP based method obtain good link recommendation performance.

\subsection{Analysis and Discussion}~\\

Figure~\ref{fig:yagoprediction} shows that the link prediction model performance widely varies from predicate to predicate in the YAGO2 knowledge base. 
For example, the HR of predicate ``dealsWith" is significantly better than ``own". Thus it is critical that we quantitatively understand the model performance
across various relations in a knowledge graph.  Recall from the \textbf{Problem Statement} that given a predicate $p$, our model $M_p$ only accounts for the bipartite subgraph $G_p$.
Motivated by~\cite{Liben.Kleinberg:03}, we study the impact of resultant graph structure of $G_p$ on the performance of $M_p$.

For each predicate $p$, we compute several graph topology metrics on its bipartite subgraph $G_p$ 
such as graph density, graph average degree, and clustering coefficient. 
Figure~\ref{fig:topology} shows the quantitative analysis between graph structure and 
link prediction model performance of each predicate. In each subfigure, x-axis represents the computed graph topology metric value of each predicate and
y-axis denotes our proposed link prediction model performance in terms of HR, ARHR, and AUC.  Each cross point shown in blue represents one specific
YAGO2 predicate used in our experiments. Then we developed a linear regression model to understand the correlation 
between link prediction model performance and each graph metric. For each linear regression curve shown in red color,
we also report its slope, intercept, and correlation coefficient (rvalue) to capture the association trend. 

From Figure~\ref{fig:topology}, both graph density and graph average degree show strong positive correlation signal with 
proposed link prediction model as demonstrated by rvalue. As our approach is inspired by collaborative filtering for recommender systems that accept a user-item matrix as input, for resultant graph of each predicate, 
higher graph density indicates higher matrix density in user-item matrix, which naturally leads to better recommendation performance in 
recommender system domain. Similar explanation can be adapted to graph average degree. For the clustering
coefficient, it shows strong negative correlation signal with link prediction model performance. For instance, in terms of AUC, the rvalue is around
$-0.69$. As clustering coefficient (cc) is the number of closed triples over the total number of triples in graph, smaller value of cc indicates
lower fraction of closed triples in the graph. Based on the transitivity property of a social graph, which states the friends of your friend have high likelihood
to be friends themselves~\cite{Zhang.Saha.ea:14, Saha.Zhang.ea:15}, it is relatively easier for link prediction model to predict (i.e.,hit) such link 
with open triple property in the graph, which leads to better link prediction performance. 

\section{Conclusion and Future Work}
Inspired by the success of collaborative filtering algorithms for recommender systems, we propose a latent feature based embedding model for the task of link prediction in a knowledge graph.  Our proposed method provides a measure of ``confidence" for adding a triple into the knowledge graph. We evaluate our implementation on the well known YAGO2 knowledge graph. The experiments show that our Bayesian Personalized Ranking based latent feature embedding approach achieves better performance compared with
two state-of-art recommender system models: Most Popular and Matrix Factorization. 
We also develop a linear regression model to quantitatively study the correlation between the performance of link prediction model itself 
and various topological metrics of the graph from which the models are constructed. The regression analysis shows strong correlation between the link prediction performance and graph topological features, such as graph density, average degree and clustering coefficient.

For a given predicate, we build link prediction models solely based on the bipartite subgraph of the original knowledge graph.  
However, as real-world experience suggests, the existence of a relation between two entities can also be predicted from the presence of other relations, 
either direct or through common neighbors. As an example, the knowledge of where someone studies and who they are friends with is useful 
to predict possible workplaces. Incorporating such intuition as ``social signals" into our current model will be the prime candidate for an immediate future work.
Another future work would be to update the knowledge graph based on the newer facts that become available over time in streaming data sources.

\section*{Acknowledgement}
This work was supported by the Analysis In Motion Initiative at Pacific Northwest National Laboratory, which is operated by Battelle Memorial Institute, 
and by Mohammad Al Hasan's NSF CAREER Award (IIS-1149851).

\bibliographystyle{abbrv}
\bibliography{link_prediction}

\balance
\end{document}